# Using crowdsourcing system for creating site-specific statistical machine translation engine

**Alexander Kalinin**
Prism LTD / Krasnoyarsk
SibSAU / Krasnoyarsk
`verbalab@yandex.ru`

**George Savchenko**
Prism LTD / Krasnoyarsk
`georgiy.savchenko`
`@langprism.com`

## Abstract

A crowdsourcing translation approach is a good tool for globalization of site content, but it is also a good source of parallel linguistic data. For the given site, processed with a crowdsourcing system, a sentence-aligned corpus can be fetched, which covers a very narrow domain of terminology and language patterns - a site-specific domain. These data can be used for training and estimation of site-specific SMT engine.

## 1   Crowdsourcing in site-translation

The more internet develops and becomes more internationalized the more question of speed of translation of web-content arises. One possible solution for increasing the speed preserving necessary level of quality is crowdsourcing translation.

According to crowdsourcing approach, the translation is performed not by a team of hired professional translators, but by a volunteering community. A well-known example of crowdsourcing project is Wikipedia, a huge web encyclopedia where articles are written and revised by anyone who is willing to contributed to the development of it. Because of this "openness" Wikipedia is developing and refreshes its data very quickly, and no maintenance from Wikipedia administrators is required to manage the project.

The same idea is behind the crowdsourcing translation: the task of translation of the content published before is distributed among members of the community. One of such translation community has appeared around on-line game "Pottermore" (http://www.pottermore.com). Users from Russian took part in translating content from the game using LangPrism tool [7], an extension for Internet browser that enables user to make translation of any web-page and to share the result of his translation.

One of the main problems of translating large web-project is the fact that new web pages demanding translation always appear, and it takes some time before translators will notice them and translate. This crowdsourcing translation gap can be filled by high quality machine translation system that is tailored to the peculiar web site. For example, it can be trained using parallel texts fetched from crowdsourcing translation activity.

## 2   Problems in site-specific statistical machine translation (SMT)

From SMT theory, it is known that a translation engine performs best if it has been trained on samples that are close to the sentences it aims to translate. Thus, SMT engine trained on corpora retrieved from a particular site will make better translations for this very site than the engine trained on any other corpus. It works because the content of a particular site has lexical and structural homogeneity.

However, site's corpus alone can't guarantee a good performance because of data sparsity — the average volume of translated sentences per site is about 7000 sentences (according to LangPrism [7] statistics). While more or less acceptable automatic translation requires at least 100k sentences for translation models and 500k for language models [1]. One possible solution for dealing with the corpus sparsity is to combine site-specific corpus with a bigger non-specified (common) corpus. Following this assumption, a site-specific translation engine to translate an on-line game was trained.

## 3 Training translation site-specific translation model with combined corpus

The site-specific corpus was obtained from www.pottermore.com on-line game translations performed by Russian community via LangPrism platform. The volume of site-specific corpus included 7k samples. It contained names of characters, places, terms of magician items used in games. The common 1M corpus was provided by Yandex [6] consisting of newspapers'. The sentences had already been 'sentence-to-sentence' aligned, so no additional sentential alignment was performed.

As an SMT system an open-source solution "Moses"[2] was used. The translation direction was from English to Russian, as the most popular one within the LangPrism project. Two corpuses were combined to shape a training set for Moses. Then, the 5-th order language model was trained on the Russian part of the corpus. After that, the translation model was trained. Phrase-based approach for training translation model was used. No other factored training was made. After the training, tuned the decoder's *.ini file using WER-metrics from MOSES was tuned. Samples for tuning were taken from the target site and did not overlap with training set. The workflow of the training is presented in Figure. 1.

To see if the quality of SMT engine had increased after some additional work, done by the translation community (some more text from the site has been translated) four iterations were performed with different specific corpus sizes.

Figure 1. Combined corpus training workflow

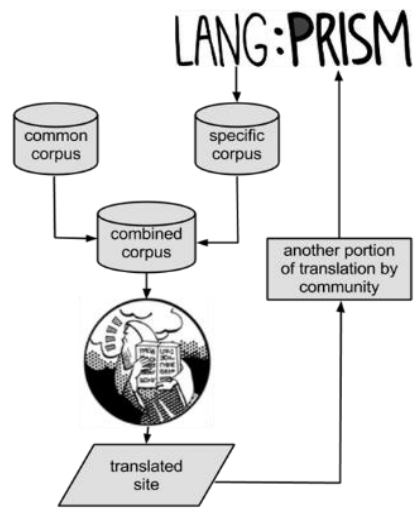

## 4 Results

Three stages of testing were conducted
1) Testing of different site-specific engines using automated metrics
2) Comparing the best site-specific engine with other MT web-services using automated metrics
3) Human pairwise estimation between best site-specific engine and best MT web-service

The testing was performed over 700 samples extracted from the site corpus, which did not appear neither in training nor in tuning samples. The performance of SMT engine was estimated with BLEU [4] and TER [5] metrics using tools from 'Moses' suite.

|  | BLEU | TER |
|---|---|---|
| Baseline (7k pottermore) | 20.10 | 0.663 |
| Yandex 1M | 10.65 | 0.764 |
| Yandex 1M + 5k pottermore | 21.49 | 0.602 |
| Yandex 1M + 6k pottermore | 22.67 | 0.643 |
| Yandex 1M + 7k pottermore | 23.78 | 0.625 |

Table 1. Comparison of different custom SMT engines trained

As it can be seen the quality of the engine increases with raises of both common corpus and specific corpus. The highest results were obtained with maximum corpora sizes.

After that, the best engine, trained on 7k specific corpus sentences and 1M common corpus sentences, was compared against other popular MT services. The test set and metrics remained the same. As it can be seen from table #2 our system outperformed all other participants of the test.

|  | BLEU | TER |
| --- | --- | --- |
| Yandex 1M + 7k pottermore | 23.78 | 0.625 |
| GoogleTranslator | 19.35 | 0.646 |
| YandexTranslator | 13.51 | 0.704 |
| MS Translator | 15.43 | 0.676 |
| Promt | 13.54 | 0.729 |

Table 2. Comparison of the best custom engine with other MT engines

In spite of higher scores in BLEU and TER it can't be stated that custom SMT engine is better than others because this metrics are automated. Moreover, in spite of correlation with human judgments the BLEU score is often under critics and is considered not very reliable by some MT expersts [4]. So it was decided to perform pairwise human estimation using method described in [3]. The estimators were invited from Russian community of www.pottermore.com, so all of them were aware of terms used in the game. Being given an original sentence, the experts were asked to choose the best from two variants or to say that they were equal in quality. In case of preference, one SMT system was scored with 1 point, in case of tie each system got 0.5 point.

380 samples' estimations were obtained. The results are the following: the system got 195.5/380 points; Google was scored 184.5/380. Therefore, the quality of the system was a little bit higher though it was within statistical error range.

## 5. Conclusion and further work

As a result, one can conclude that training custom statistical machine translation engine using two corpora – a common one and specific one – can lead to raise of quality comparing with training using specific or common corpus only.

Moreover, a custom engine trained on rather modest corpus (1M+7k) using open-source solution can compete with and even outperform popular translation system from companies like Google and MS. In addition, this level of quality is obtained without any sophisticated preprocessing and linguistic analysis, using just the data from crowdsourcing translation. Of course, such quality can be obtained only for the site, which the system was designed to translate, and it would show much poorer results on other sites.

The problem of system was that it showed quite poor quality with long sentences while Google worked well with them. That happened because our system did not use any linguistic information to build good sentences, that's was one of the fact why it did not get much points in manual estimation where Google translator produced good grammatical sentences. Therefore, some deeper training factoring linguistic issues is necessary for building a custom site-specific SMT system more preferable to large system like Google and MS.